\documentclass{article}
\usepackage{spconf,amsmath,graphicx}
\usepackage{epstopdf}
\usepackage{mathtools, xparse}
\usepackage{bbm}
\usepackage{multirow}
\usepackage{array}
\usepackage[lofdepth,lotdepth]{subfig}
\usepackage[normalem]{ulem}
\usepackage{amsmath,graphicx}
\usepackage{epstopdf}
\usepackage{epsfig}
\usepackage{times}
\usepackage{epsfig}
\usepackage{graphicx}
\usepackage{amssymb}
\usepackage[]{algorithm}
\usepackage[]{algorithmic}
\usepackage{tabularx}
\usepackage{adjustbox}
\usepackage{accents}
\usepackage{color}
\usepackage{url}
\usepackage[table]{xcolor}
\usepackage{pbox}
\usepackage{booktabs}
\usepackage{comment}

\usepackage{color}

\DeclareMathOperator*{\argmin}{argmin}

\title{QUADRUPLET SELECTION METHODS FOR DEEP EMBEDDING LEARNING}

\name{Kaan Karaman$^{1,4}$, Erhan Gundogdu$^{2}$ $^{\dagger}$, Aykut Ko\c{c}$^{1}$, A. Ayd\i n Alatan$^{3,4}$}
\address{$^1$Aselsan Research Center, Ankara, Turkey $^2$ CVLab, EPFL, Switzerland\\ $^3$Center for Image Analysis (OGAM) $^4$Electrical and Electronics Eng. Dept., 					METU
		Ankara, Turkey\\
		\{kkaraman, aykutkoc\}\{at\}aselsan.com.tr, erhan.gundogdu\{at\}epfl.ch, alatan\{at\}metu.edu.tr}

\begin{document}
%\ninept

\maketitle

\begin{abstract}
Recognition of objects with subtle differences has been used in many practical applications, such as car model recognition and maritime vessel identification. For discrimination of the objects in fine-grained detail, we focus on deep embedding learning by using a multi-task learning framework, in which the hierarchical labels (coarse and fine labels) of the samples are utilized both for classification and a quadruplet-based loss function. In order to improve the recognition strength of the learned features, we present a novel feature selection method specifically designed for four training samples of a quadruplet. By experiments, it is observed that the selection of very hard negative samples with relatively easy positive ones from the same coarse and fine classes significantly increases some performance metrics in a fine-grained dataset when compared to selecting the quadruplet samples randomly. The feature embedding learned by the proposed method achieves favorable performance against its state-of-the-art counterparts.

%\textit{Recall@K} and Normalized Mutual Information (\textit{NMI}) scores

\end{abstract}


{\let\thefootnote\relax\footnote{{$^{\dagger}$ This work was done when Erhan Gundogdu was with Middle East Technical University.}}}

%{\let\thefootnote\relax\footnote{{This  work  has  been  accepted  by  2019  26$^{th}$ IEEE International Conference on Image Processing (ICIP 2019).}}}

{\let\thefootnote\relax\footnote{{Copyright 2019 IEEE. Published in the IEEE 2019 International Conference on Image Processing (ICIP 2019), scheduled for 22-25 September 2019 in Taipei, Taiwan. Personal use of this material is permitted. However, permission to reprint/republish this material for advertising or promotional purposes or for creating new collective works for resale or redistribution to servers or lists, or to reuse any copyrighted component of this work in other works, must be obtained from the IEEE. Contact:Manager, Copyrights and Permissions / IEEE Service Center / 445 Hoes Lane / P.O. Box 1331 / Piscataway, NJ 08855-1331, USA. Telephone: + Intl. 908-562-3966.}}}

\section{Introduction}
\label{sec:intro}

Recently, embedding learning has become one of the most popular issues in machine learning \cite{smartMining, proxyMethod, clusteringEmbedding}. Proper mapping from the raw data to a feature space is commonly utilized for image retrieval \cite{NPairLoss} and duplicate detection \cite{adversarialStabilityTraining}, which are used in many applications such as online image search. 

For training a model that can extract proper features, the distance between two samples of a dataset in the feature space should be considered. Moreover, some embedding learning methods are employed to increase the classification accuracy, \emph{e.g.}, fine-grained object recognition \cite{labelStructure} by using deep convolutional neural network (CNN) models which require a significant amount of training samples. Fortunately, there are datasets for various purposes such as car model recognition \cite{stanfordCarsDataset} and maritime vessel classification and identification \cite{marvelDataset}. Some of these datasets can be used for classifying land, marine, and air vehicles in a real-world scenario. Concretely, car model recognition can be employed in the context of visual surveillance and security for the land traffic control \cite{labelStructure} and marine vessel recognition is used for the purpose of coastal surveillance \cite{SIUErhan} \cite{GenericBerkan}. In this work, we focus on the feature learning problem specifically designed for car model recognition.

Recently developed studies on feature learning focus on extracting features from raw data such that the samples belonging to different classes are well-separated and the ones from the same classes are close to each other in the feature space. The state-of-the-art network architectures such as \textit{VGG} \cite{vggNetwork} and \textit{GoogLeNet} \cite{goingDeeper} are frequently used for extracting features from images by several different training processes. In the early years, pairwise similarity is used for signature verification with \textit{contrastive} loss \cite{siamese}. Since consideration of the whole pairs or triplet samples in a dataset is not computationally tractable, carefully designed mining techniques are proposed, such as hard positive \cite{discriminativeHardPositive} and negative \cite{unsupervisedHardNegative} mining.

In the previous methods that employ a hard mining step during training, at each iteration of the optimization, they focus on the separation of samples in the feature space in a selected batch from the dataset. Therefore, the distance relations among the samples in a dataset are not fully exploited. Moreover, the classification loss function for the fine-grained labels is not considered in the training phase. On the other hand, our proposed method for the quadruplet sample selection enables to convey more information from the utilized dataset by considering the globally hard negatives and relatively easy positives in the distance loss terms and the auxiliary classification layers.

The contributions of this work are summarized as follows: (1) In order to improve embedding learning, we have proposed two novel quadruplet selection methods where the globally hardest negative and moderately easy positive samples are selected. (2) Our framework contains a CNN trained with the combination of the classification and distance losses. These losses are designed to exploit the hierarchical labels of the training samples. (3) To test the proposed method, we have conducted experiments on the \textit{Stanford Cars 196} dataset \cite{stanfordCarsDataset} and observed that the recognition accuracy of the unobserved classes has been improved with respect to the random selection of samples in the quadruplets while outperforming the state-of-the-art feature learning methods.

\section{Related Work}
\label{sec:related}

Earlier works on metric learning are based on \textit{Siamese Nets} \cite{siamese}. In that study, two identical neural networks extract the features of two arbitrary images. Next, these features are compared by a metric which is based on a radial function\footnote{The distance between any two members in the feature space is defined as the cosine of the angle between them \cite{siamese}. }. While their loss function forces the samples in the same class to be closer to each other in the sense of the selected distance function, the samples in the different classes are forced to be mapped far from each other. The cost function of such a network is given below \cite{constrastiveLoss} where $[.] _ +$ represents the operation of $max(0,.)$, and $D_{i, j}$ are distances in between samples.
\begin{equation}
\label{siameseLoss}
\mathcal{L}_{siamese}(i,j) = y_{i,j}D_{i,j}^2 + (1-y_{i,j})[\alpha-D_{i,j}]_+^2.
\end{equation}

A similar approach uses triplets for training process as in \cite{tripletDistance}, where each triplet sample consists of three members: (1) Reference (anchor) sample, $X^R$, (2) Positive sample, $X^P$, (3) Negative sample, $X^N$. The constraints of a triplet are as follows: the reference and positive samples belong to the same class, whereas the negative sample does not ($ X^R \in C_i $, $ X^P \in C_i $, and $ X^N \not \in C_i $, where $C_i$ denotes the class label of the reference sample). For well-separation of the classes, $X^R$ should be closer to $X^P$ than $X^N$. The selection method of triplets is known to be an important issue for convergence \cite{tripletDistance}. Among the existing studies, some of them indicate that selecting the samples randomly reduces the efficiency of training. A recent study in \cite{unsupervisedHardNegative} proposes hard negative mining, which emphasizes that selecting $X^N$ close to $X^R$ increases the performance of separation in the feature space. On the other hand, hard positive mining is also suggested to enhance the performance by selecting $X^P$ far from $X^R$ \cite{discriminativeHardPositive}. Moreover, hard negative and positive mining methods are also used for the face recognition purpose \cite{faceNet}. For triplet-based approaches \cite{tripletLossFunction}, the following function is utilized where the distances are defined as $l_2$ norm\footnote{The distance between any two members ($x_i$ and $x_j$) in the space is defined as $ D_{ij} = || f_\theta(x_i) - f_\theta(x_j) ||_2 $.}, and $m$ is a margin:
\begin{equation}
\label{tripletloss1}
\mathcal{L}_{triplet}(x^R,x^P,x^N,m)=[D_{R,P}^2-D_{R,N}^2+m]_+.
\end{equation}

Another approach is to utilize the hierarchical class labels of the training samples \cite{labelStructure}. In that method, samples with similar fine labels have the same coarse label, i.e. a sample has more than one label. The cost function is modified by considering both the coarse and fine labels. For this purpose, each quadruplet sample is constructed as follows: (1) Reference sample (anchor sample), $X^R$, (2) Positive positive sample, $X^{P^+}$, (3) Positive negative sample, $X^{P^-}$, (4) Negative sample, $X^N$. Similar to the triplet selection, the quadruplets are selected such that three constraints should be taken into account. First, both the coarse and fine classes of $X^R$ and $X^{P^+}$ should be the same. Second, although the coarse class of $X^R$ is the same as the coarse class of $X^{P^-}$, the fine classes are different. Finally, the coarse class of $X^R$ and $X^N$ should be different. 

%In a formal way, the constraints for the quadruplets where $y$ is a quadruplet sample (consists of four samples) are given in Equation \eqref{constraintQuad}. $y_{i1}$ and $y_{i2}$ denote the coarse and fine classes in the equation, respectively.
%\begin{equation}
%\begin{split}
%\label{constraintQuad}
%y_{i1}^R = y_{i1}^{P^+} = y_{i1}^{P^-} \neq y_{i1}^{N}, \\
%y_{i2}^R = y_{i2}^{P^+} \neq y_{i2}^{P^-}. 
%\end{split}
%\end{equation}

Moreover, the loss function for the quadruplets is similar to the triplet based methods \cite{labelStructure}. On the other hand, in \cite{SIUErhan}, the use of the global loss has been proposed, while the quadruplet samples are selected randomly (Note that these quadruplets hold the constraints). The global loss penalizes the network in case of the mean and variance of the distances between the samples in a quadruplet are not appropriate, as given in \eqref{globalLoss}\footnote{In \eqref{globalLoss}, $\sigma^2_{P^{+/-}} = var\{D_{R, P^{+/-}}\}$, $\sigma^2_{N} = var\{D_{R, N}\}$, and $\mu_{P^{+/-}} = E\{D_{R, P^{+/-}}\}$, $\mu_{N} = E\{D_{R, N}\}$ as defined in \cite{triplet+global}.}, where $t_1$ and $t_2$ are the margins, similar to \eqref{tripletloss1}.
\begin{equation}
\begin{split}
\label{globalLoss}
\mathcal{L}_{global}(Q) = \sigma^2_{P^+} &+ \sigma^2_{P^-} + \sigma^2_{N} + \\
\lambda_{g1} [\mu_{P^+} - \mu_{P^-} + t_1 - t_2 ]_+ &+ \lambda_{g2} [\mu_{P^-} - \mu_{N} + t_2 ]_+.
\end{split}
\end{equation}

In \cite{labelStructure}, the hierarchical labels of the training samples are utilized. It should be noted that a model has difficulty in convergence when the samples are selected randomly since the most informative pairs are not effectively considered. Here, we propose two methods for sample selection to address this issue.
%\vspace{-2mm}

\section{Proposed Method}
\label{sec:method}

Each quadruplet sample is represented as $Q_i = \{X_i^R, X_i^{P^+},$ $ X_i^{P^-}, X_i^{N} \}$ where $X_i = (x_i, y_{i1}, y_{i2}) $. $ x_i \in \mathcal{R}^n$ represents the vector of the pixels of an image ($n$ is the number of the pixels in the image), $y_{i1} \in C_1$ and $y_{i2} \in C_2$ represents the coarse, and fine classes, respectively, where $C_1 = \{c^i_1\}_{i=1}^{k_1}$ ($k_1$ is the number of coarse classes) and similarly, $C_2 = \{c^i_2\}_{i=1}^{k_2}$. Let the weights of a CNN be $\theta \in \mathcal{R}^m$ where $m$ is the number of the weights, then the network can be defined as $f_\theta(x_i) : \mathcal{R}^m \times \mathcal{R}^n \rightarrow \ \mathcal{R}^k$ where $k$ is the dimension of the feature space.

Our proposed cost function consists of two parts: the classification (Section \ref{ssec:classCost}) and distance (Section \ref{ssec:distCost}) cost functions. The aim of these cost functions is to form the feature space so that fine classes are well-separated. However, the learning process highly depends on the selection of the quadruplets. The training process takes more time when selecting the quadruplets in an erroneous strategy. We propose to select the members of the quadruplets from the most informative region in the feature space in Section \ref{ssec:quad}. As validated by the experiments (Section \ref{sec:results}), proposed method increases the performance of separation significantly as it can be observed from both \textit{Recall@K} and Normalized Mutual Information (\textit{NMI}) values in Table \ref{verificationTable}.

\subsection{Classification Cost Function}
\label{ssec:classCost}

In order to increase the discriminativeness of the features for the available class labels, \textit{softmax} loss is employed. Contrary to the traditional one, the proposed neural network has two outputs which are dedicated to the fine and coarse classes. Let $s_\theta = [g_\theta, h_\theta ]$ where $g_\theta$ denotes the output for the coarse class, whereas $h_\theta$ is for the fine class. Then, the proposed cost function is obtained:
\begin{equation}
\begin{split}
\label{softmaxloss}
\mathcal{L}_{C1,C2}(x) = -\lambda_{c1}\sum_{i=1}^{k_1} p(c^i_1) log 
\left( \frac{e^{h_\theta^x(c^i_1)}}
{\sum_{j=1}^{k_1} e^{h_\theta^x(c^j_1)}}\right) \\
-\lambda_{c2} \sum_{i=1}^{k_2} p(c^i_2) log 
\left( \frac{e^{g_\theta^x(c^i_2)}}
{\sum_{j=1}^{k_2} e^{g_\theta^x(c^j_2)}}\right).
\end{split}
\end{equation}

$C_1$ and $C_2$ specify the coarse and fine classes, respectively. $p(c^i_1)$ is the probability that the $x$ vector belongs to the $i^{th}$ coarse class. If $x \in c^j_1$, then by using hard decision, $p(c^i_1) = \delta_{ij}$ where $\delta_{ij}$ is the Kronecker delta function. Similarly, $p(c^i_2)$ is also calculated for $C_2$. $h_\theta^x(c^i_1)$ represents the $i^{th}$ element of the $h_\theta^x$ vector, where $h_\theta^x$ is the score vector for the coarse classes ($C_1$). Likewise, $g_\theta^x$ is the one for the fine classes ($C_2$). $\lambda_{c1}$ and $\lambda_{c2}$ are the weights of the fine and coarse classification terms of the cost function.

\subsection{Distance Cost Function}
\label{ssec:distCost}

The distances between the samples in the feature space are commonly defined by a radial function \cite{tripletDistance}. For this reason, the representations which will be learned by our proposed framework are $m$-dimensional feature vectors. The distance for any two members can be defined by $l_2$ norm. Hence, we can clearly formulate our goal by the inequality $ D_{R,P^+} < D_{R,P^-} < D_{R,N} $. The first part can be rewritten as $D_{R,P^+} + m_1 < D_{R,P^-}$, and the second part would be $D_{R,P^-} + m_2 < D_{R,N}$ where $m_1$ and $m_2$ are the margins, which should be positive numbers. Moreover, we emphasize the discrimination of the coarse classes by using the condition $m_1 > m_2 > 0$. Then, the new cost function can be proposed as:
\begin{equation}
\label{jointLoss}
\begin{gathered}
\mathcal{L}_{joint}(x^R,x^{P^+},x^{P^-},x^N) = {\left[1-\frac{D_{R,P^-}}{D_{R,P^+}+m_1-m_2}\right]}_+ \\
+\left[1-\frac{D_{R,N}}{D_{R,P^-}+m_2}\right]_+ + \mathcal{L}_{C1,C2}(x^R).
\end{gathered}
\end{equation}

\begin{figure}[ht]
\centering
%\shorthandoff{=}
\includegraphics[width=.42\textwidth]{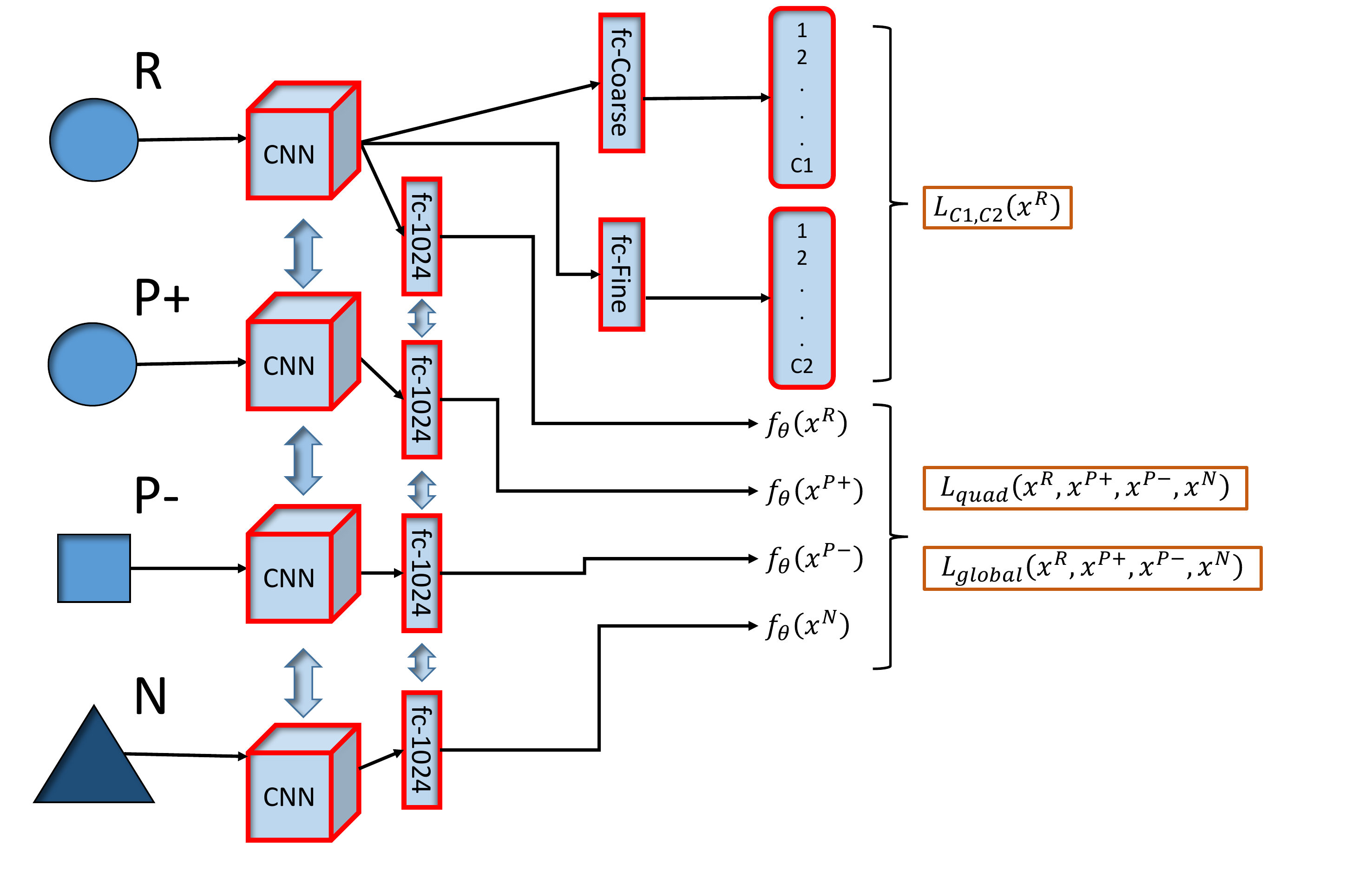}
\caption{The proposed framework is similar to the model used in \cite{SIUErhan}. The dimension of the last fully connected (FC) layer is $1024$. Note that all the weights in the network are shared, including the weights in the FC layers.}
\label{fig:QuadNetwork}
\normalsize
\end{figure}

Finally, the overall proposed network is shown in Figure \ref{fig:QuadNetwork} with the loss function given in \eqref{combLoss}. This loss function, which is the combination of \eqref{jointLoss} and \eqref{globalLoss}, consider the distances of the samples in the feature space using $\mathcal{L}_{joint}$ while $\mathcal{L}_{global}$ regularizes the statistics of the distances batch-wise.

\begin{equation}
\label{combLoss}
\begin{gathered}
\mathcal{L}_{comb}(Q) = \sum_{\forall i} \mathcal{L}_{joint}(Q_i) + 
\eta \mathcal{L}_{global}(Q).
\end{gathered}
\end{equation}

%\vspace{-5mm}

\subsection{Quadruplet Selection}
\label{ssec:quad}

In the previous section, we have briefly summarized our novel loss function. As it is mentioned before, selecting the quadruplet samples randomly makes it difficult to exploit the most informative training examples. Instead of attempting to cover all the quadruplet combinations in the training set, we propose two novel selection strategies. First, a reference sample is randomly selected with equal probability from the training set (Let the reference sample be selected as $X^R$, where $C_1^R$ and $C_2^R$ are the coarse and fine labels of the reference sample, respectively.). The negative sample is selected from the set of the samples belonging to the different coarse classes. The critical point is that, like hard negative mining in \cite{unsupervisedHardNegative}, we should select the closest negative sample to $X^R$ ($X^N := \argmin\limits_{X^N \not\in C_1^R} || f_\theta(x^R) - f_\theta(x^N) ||_2$). At this point, we propose two different methods for the selection of $X^{P^+}$ and $X^{P^-}$. The experimental comparison of these two methods is given in Section \ref{sec:results}.
%\vspace{-3mm}

\subsubsection{Method 1}
\label{sssec:method1}
For determining $X^{P^+}$, we select the sample whose fine class is the same as the fine class of $X^R$, and which is closest to $X^N$. At this point, the constraint for selection of $X^{P^+}$ is as follows: the distance between $X^{P^+}$ and $X^R$ is greater than the distance between $X^R$ and $X^N$ ($D_{R,P^+} > D_{R,N}$). Similarly, we select $X^{P^-}$ whose coarse class is the same as the coarse class of $X^R$, which is the closest sample to $X^N$, and also satisfying $ D_{R,P^-} > D_{R,N} $. This method is visualized in Figure \ref{fig:methods}.

%\vspace{-2mm}

\subsubsection{Method 2}
\label{sssec:method2}

\begin{figure}[ht]
\centering
%\shorthandoff{=}
\includegraphics[width=.40\textwidth]{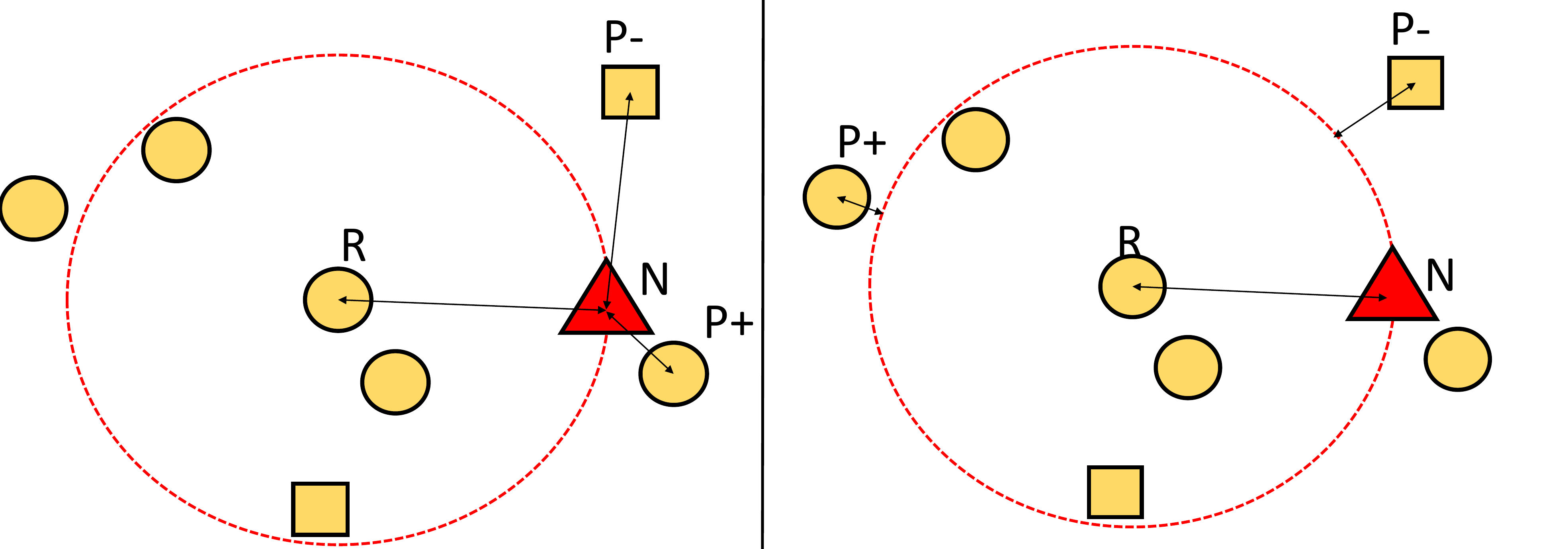}
\caption{The two-dimensional representation of the samples in the feature space. The different shapes represent the different fine classes, the different colors represent the different coarse classes. After $X^R$ is selected, the nearest sample belonging to the different coarse class is selected as $X^N$. $X^{P^+}$ and $X^{P^-}$ are also selected as in \textit{Method} $1$ (left), and \textit{Method} $2$ (right).}
\label{fig:methods}
\normalsize
\end{figure}

In the second method, after selecting $X^N$, the distance between $X^R$ and $X^N$ ($D_{R,N}$) determines a hyper-sphere which takes $X^R$ as its center. After selecting the labels of $X^{P^+}$ and $X^{P^-}$ according to the constraints in Section \ref{sec:related}, $X^{P^+}$ and $X^{P^-}$ are selected from the predetermined classes such that they are the closest points to $X^R$ but outside the region enclosed by this hyper-sphere. If there are no samples which are both close to $X^R$ and outside of the hyper-sphere, then the furthest sample to $X^R$ inside the hyper-sphere is selected. This selection method is illustrated in Figure \ref{fig:methods}.

%\vspace{-4mm}

\section{Results}
\label{sec:results}

We compare the performance of our proposed method against the state-of-the-art feature learning approaches in \cite{faceNet, liftedStructure, NPairLoss, clusteringEmbedding, triplet+global} by using the same evaluation methods. In addition, the randomly selected quadruplets are utilized as in \cite {SIUErhan}. \textit{Stanford Cars 196} dataset \cite{stanfordCarsDataset} is used in the experiments. To implement the proposed methods, a hierarchical structure is required for all the samples in the dataset, where each sample originally has only one label. For this purpose, we should add the high-level classes (coarse labels) to the dataset. In other words, the $196$ classes, which are originally in the dataset, are taken as the fine classes and $22$ coarse classes are added using the types of the cars, similar to the study in \cite {labelStructure}. 

The important point in the generation of the training and test sets is that they should not share any fine class labels. With this restriction, we want to measure the adequacy of our neural network to separate the classes that have not been seen before. The most common performance analysis methods for zero-shot learning are \textit{Recall@K} and \textit{NMI}. \textit{Recall@K} specifies whether the samples belonging to the same fine class are close to each other, and \textit{NMI} is a measure of clustering quantity as mentioned in \cite{clusteringEmbedding}.

For this purpose, the first $98$ fine classes of the dataset are selected as the training set, and the rest are used only as the test set similar to the study in \cite{smartMining}. In our experimental setup, the pre-trained \textit{ResNet101} model \cite{resnet} (that has been trained using the \textit{ImageNet} dataset \cite{imageNet2015}) is employed as our CNN model to extract the features. The experiments are performed on \textit{Pytorch} platform \cite{pytorch}. In addition, the hyper-parameters of the cost function are selected as $ 0.08 $ for $ \lambda_{c1} $, $ 0.25 $ for $ \lambda_{c2} $; $1$ for $\lambda_{g1}$, $\lambda_{g2}$, and $\eta$. The margins are $ 0.7 $ for $ m_1 $, and $t_1$; $ 0.3 $ for $ m_2 $, and $t_2$. The learning parameters are as follows: the learning rate is $ 0.0003 $, the momentum is $ 0.9 $, and \textit{stochastic gradient descent} algorithm is used for optimization. The results can be examined in Table \ref{verificationTable}.

\begin{table}[ht]
\centering
\resizebox{.48\textwidth}{!}{%
\begin{tabular}{|c|c|c|c|c|c|}
\hline
\textbf{Method} & \textbf{R@1} & \textbf{R@2} & \textbf{R@4} & \textbf{R@8} & \textbf{NMI}\\
\hline
Semi-Hard \cite{faceNet} & 51.54 & 63.78 & 73.52 & 82.41 & 55.38\\
\hline
Lifted Structure \cite{liftedStructure} & 52.98 & 65.70 & 76.01 & 84.27 & 56.50\\
\hline
N-Pairs \cite{NPairLoss} & 53.90 & 66.76 & 77.75 & 86.35 & 57.24\\
\hline
Clustering \cite{clusteringEmbedding} & 58.11 & 70.64 & 80.27 & 87.81 & 59.23\\
\hline
Triplet $+$ Global \cite{triplet+global} & 61.41 & 72.51 & 81.75 & 88.39 & 58.61\\
\hline
Random Quadruplet Selection \cite{SIUErhan} & 61.49 & 73.41 & 82.88 & 89.92 & 54.50\\
\hline
Proposed Method 1 & \underline{64.85} & \underline{75.59} & \underline{83.41} & \underline{89.55} & \underline{57.32}\\
\hline
Proposed Method 2 & \textbf{66.06} & \textbf{76.62} & \textbf{84.84} & \textbf{90.63} & \textbf{57.00}\\
\hline
\end{tabular}}
%\vspace{2mm}
\caption{Using \textit{Stanford Cars 196} dataset, precision of \textit{Recall@K} and \textit{NMI} are shown for different methods.}
\label{verificationTable}
\end{table}

Our proposed quadruplet based learning framework has improved the precision in terms of \textit{Recall@K} even if they are selected randomly. According to \textit{Recall@K} metric, random quadruplet selection method outperforms the previous studies in \cite{faceNet, liftedStructure, NPairLoss, clusteringEmbedding}, and it is comparable to the study in \cite{triplet+global}. On top of that, when the proposed selection methods are used, even higher levels of accuracy can be obtained. As it is demonstrated in Table \ref{verificationTable}, \textit{Method} $1$ results in $ 64.85\% $ accuracy of \textit{Recall@1}, which is an improvement by at least $3.4\% $ compared to the other studies; while \textit{Method} $2$ results in $ 66.06\% $ accuracy of \textit{Recall@1} corresponding to a $ 4.5\% $ increase.

\section{Conclusion}
\label{sec:conclusion}
We have demonstrated the proposed method of selection significantly increases the rate of separation of a model in terms of recall performance. Unlike previous studies that consider only the distances between $X^R$-$X^{P^{+/-}}$ and $X^R$-$X^N$, the proposed methods consider also the distances between $X^N$- $X^{P^{+/-}}$ in the feature space. This consideration helps us improve the model and achieve better accuracy performance. These two proposed selection methods allow the loss function not only to enlarge margins between the samples in the different classes but also to create several tight clusters for each class. Moreover, these two proposed methods have the advantage that they pay attention to the samples at the region around the critical hyper-sphere. Especially, the second method attacks the easier problem, i.e. while the first method can reshape the only particular region in the feature space, the second one can use all the region on the surface of a hyper-sphere. Therefore, the feature space is manipulated through a better optimization procedure. 

%\clearpage
\bibliographystyle{IEEEbib}
\bibliography{refs}

\end{document}